\documentclass[sigconf, authordraft]{acmart}

\usepackage{subfigure}
\usepackage{graphicx}
\usepackage{amsmath}
\usepackage{bm}
\usepackage{url}
\usepackage{stmaryrd}
\usepackage{balance}
\usepackage{amssymb}
\usepackage{pgfplots}
\usepackage{pifont}
\usepackage{epsfig}
\usepackage{booktabs}
\usepackage{pgffor}
\usepackage{tikz}
\usepackage{multirow}
\usepackage{amsmath}
\usepackage{bbm}

\makeatletter
\newcommand\xleftrightarrow[2][]{%
  \ext@arrow 9999{\longleftrightarrowfill@}{#1}{#2}}
\newcommand\longleftrightarrowfill@{%
  \arrowfill@\leftarrow\relbar\rightarrow}
\makeatother

\newcommand{\mb}{\mathbf}

\newcommand{\our}{\textsc{EgoCoder}}
\newcommand{\problem}{automatic program synthesis}
\newcommand{\unit}{\textsc{Hsu}}

\newcommand{\astbilstm}{\textsc{Ast-BiRnn-LSTM}}
\newcommand{\astbigru}{\textsc{Ast-BiRnn-GRU}}
\newcommand{\astbirnn}{\textsc{Ast-BiRnn-Basic}}

\newcommand{\bilstm}{\textsc{BiRnn-LSTM}}
\newcommand{\bigru}{\textsc{BiRnn-GRU}}
\newcommand{\birnn}{\textsc{BiRnn-Basic}}

\newcommand{\autoencoder}{\textsc{Ast-AutoEncoder}}

\usepackage{algorithm}
\usepackage{algorithmic}

\usepackage{color}

\setcopyright{rightsretained}

\begin{document}

 



\title{EgoCoder: Intelligent Program Synthesis with Hierarchical Sequential Neural Network Model}

\author{Jiawei Zhang$^1$, Limeng Cui$^2$, Fisher B. Gouza$^1$}
\affiliation{%
  $^1$IFM Lab, Department of Computer Science, Florida State University, FL, USA \\
  {$^2$School of Computer and Control Engineering, University of Chinese Academy of Sciences, Beijing, China}
}
\email{jzhang@cs.fsu.edu, lmcui932@163.com, fisherbgouza@gmail.com}

\begin{abstract}

Programming has been an important skill for researchers and practitioners in computer science and other related areas. To learn basic programing skills, a long-time systematic training is usually required for beginners. According to a recent market report, the computer software market is expected to continue expanding at an accelerating speed, but the market supply of qualified software developers can hardly meet such a huge demand. In recent years, the surge of text generation research works provides the opportunities to address such a dilemma through automatic program synthesis. In this paper, we propose to make our try to solve the program synthesis problem from a data mining perspective. To address the problem, a novel generative model, namely {\our}, will be introduced in this paper. {\our} effectively parses program code into abstract syntax trees (ASTs), where the tree nodes will contain the program code/comment content and the tree structure can capture the program logic flows. Based on a new unit model called {\unit}, {\our} can effectively capture both the hierarchical and sequential patterns in the program ASTs. Extensive experiments will be done to compare {\our} with the state-of-the-art text generation methods, and the experimental results have demonstrated the effectiveness of {\our} in addressing the program synthesis problem.
\end{abstract}

\keywords{Program Synthesis; Text Generation; Neural Networks; Data Mining}


\maketitle

\section{Introduction}\label{sec:introduction}

Formally, programing denotes the process of developing and implementing computer instructions to enable a computer to perform certain tasks. These instructions are usually written in one or several programing languages, and a sequence of computer instructions (implementing the pre-specified functions) will be called a computer program, which helps the computer to operate smoothly. To learn necessary programing skills, a long-time systematic training is usually required for beginners. Generally, to be a qualified programmer, people may need to master knowledge from various areas, including \textit{programing language}, \textit{discrete mathematics}, \textit{data structure} and \textit{algorithm}, etc. 

Computer programing continues to be a necessary and important skill for both academic researchers and industry practitioners as the Internet and AI applications continue to expand. As introduced in \cite{grow}, the computer software market is expanding at an accelerating speed and is estimated to grow from $19.98$ Billion USD in 2014 to more than $50.34$ Billion USD in 2022. Meanwhile, according to the latest market analysis report \cite{market}, there exists a huge gap between the market supply and demand of software developers. For instance, from January 2016 to February 2017, more than $115,000$ job postings requesting for qualified software engineers have been posted in each month, but the average monthly hire number is merely $33,579$. Such a huge demand-supply gap also motivates many large IT companies to seek for other ways to address such a problem.

For effective program code storage and maintenance, inside all the well-known big IT and related technology companies, they are maintaining a company-internal program codebase for storing all the developed program code of company systems, web services, software products and research projects. The program code in these codebases is normally of a tremendous amount. A recent report \cite{codebase} releases the lines of code used in several companies and software systems, among which Google ranks the top with more than 2 billion lines of code \cite{google} used in all its Internet services. These company codebase repositories cover very diverse yet high-quality code, which are also the most valuable intellectual property of companies, but fail to be effectively exploited.

Programing has been long-time treated as one of the most challenging skills mastered by a very small number of people from some untrained eyes. In this paper, we will make our try to attack this holy-grail pride of software engineers by training a model to write programs automatically. The {\problem} problem is a fundamental problem from the technology, business and society development perspectives. Successfully addressing the problem will effectively bridge the market supply\&demand gap for qualified practitioners, greatly stimulate the development of IT and other related areas, intelligently recycle the company internal codebase for secondary-development, and promisingly free human from the tedious coding positions to other more challenging jobs.

In recent years, due to the surge of deep learning developments \cite{GBC16}, many text generation research works and models have been proposed, which introduce many novel yet interesting research problems. Meanwhile, slightly different from the unstructured sentences written in natural languages, the program code written in programing languages is highly structured, which can be precisely parsed into a hierarchical structure according to the specific programming language grammar. For instance, for the program written in an advanced programing language, Python, its code will consist of hierarchical structures like \textit{class}, \textit{function}, \textit{statements} and \textit{expressions}, etc. Therefore, instead of handling the program characters by characters (like the existing text generation research works \cite{SMH11}), new techniques that can handle the program according to its own structure will be necessary. 

The {\problem} problem is extremely challenging to solve due to several reasons:
\begin{itemize}

\item \textit{Lack of Problem Definition}: The {\problem} problem is still an open problem to this context so far. A formal definition of {\problem} will be required before proposing potential solutions to address it.

\item \textit{Program Hierarchical Structure Extraction}: There usually exists a concrete hierarchical-sequential structure of program code according to its logic flows hierarchically and sequentially. Generally, code tokens at the lower level of programs will precisely implement the desired physical functions of the program components at higher levels; meanwhile, at each level, the logic will flow in a sequential manner from the beginning to the end. Extraction of such a hierarchical-sequential program structure will be useful for effective program information modeling and representation learning.

\item \textit{Unit Model}: For each component in the hierarchical-sequential structure aforementioned, depending on the specific running mode, it will accept the input from the components above/below and before/after the component. A new unit model for implementing such an intertwined relationships in the learning process will be desired.

\item \textit{Program Intention Incorporation}: Besides the program code itself, there usually exist some textual descriptions of the program code in a natural language, which indicates the physical function of the program, e.g., \textit{ranking}, \textit{shuffling}, \textit{searching}, \textit{factorization} and \textit{dynamic programing}, etc. Effectively incorporating the program intention into the learning process will allow both program generation and interpretation across natural languages and programing languages.
\end{itemize}

To effectively resolve the above challenges, in this paper, we will introduce a novel neural network model, namely {\our}, with a deep architecture. {\our} provides a formal definition of the {\problem} problem, which covers three different sub-problems respectively: program generation, program interpretation and program completion. Instead of learning the models based on the pure text information in the program code, {\our} extracts the hierarchical-sequential structure with a programming language parser, which translates the input program code into abstract syntax tree (AST) structured diagrams. For each node in the extracted ASTs, it contains both syntax types and tokens as its content. Meanwhile, the structure of the extracted ASTs will also effectively indicate the semantic logic flows of the program. To capture both the syntax contents of the program components and the semantical logic flow of the program, a new unit model, namely {\unit} (hierarchical sequential unit), will be used as the basic component in {\our}. Unit model {\unit} can accept inputs from sibling nodes at the same levels, as well as evolving information from the child nodes and inheriting information from the father node simultaneously. Based on a set of sampled sub-tree batches from the extracted program ASTs, {\our} can be trained effectively to capture the substructures covered in the ASTs. These new technical terms mentioned above will be clearly illustrated in great details in this paper.


\section{Problem Formulation} \label{sec:formulation}

In this section, we will first define several important concepts used in this paper, based on which we will provide the formulation of the studied problem and its three different running modes.

\subsection{Terminology Definition}

Computer program usually has a highly structured hierarchy, involving the code components belonging to different syntax types.

\begin{definition}
(Program Syntax Type): Formally, we can represent the set of syntax types involved in the program as set $\mathcal{C} = \{$\textit{module}, \textit{class}, \textit{function}, \textit{statement}, \textit{expression}$\} \cup \{$\textit{unit token syntax type}$\}$, where the unit token syntax type set involves various variable and operator types used in the program.
\end{definition}

Based on the \textit{program syntax type} set, we can translate a program into a \textit{program abstract syntax tree}, where the nodes denote program code components (i.e., code blocks) belonging to different syntax types, and the links represent the semantic logic flows among the code components.

\begin{definition}
(Program AST): Formally, a \textit{program AST} can be represented as a graph structured diagram: $T = (\mathcal{V}, \mathcal{E}, root)$, where $\mathcal{V}$ denotes the set of program component nodes, and $\mathcal{E}$ denotes the set of \textit{logic-flow} relationships among the nodes at either different hierarchical levels or at the same hierarchical levels. In $T$, the $root \in \mathcal{V}$ represents the top program component node, which usually denotes the program \textit{module} component by default.
\end{definition}

\begin{definition}
(Program Component Node): Each \textit{program component node} $v \in \mathcal{V}$ in the \textit{program AST} can be denoted as a triple $v = (c, t, f)$, where $c \in \mathcal{C}$ denotes its syntax type, $t$ represents its textual content and $f$ denotes the functional intention of the program component. The overall program intention can be represented as the AST root node intention by default.
\end{definition}

\begin{figure}
\vspace{-30pt}
	\centering
	\includegraphics[width=0.45\textwidth]{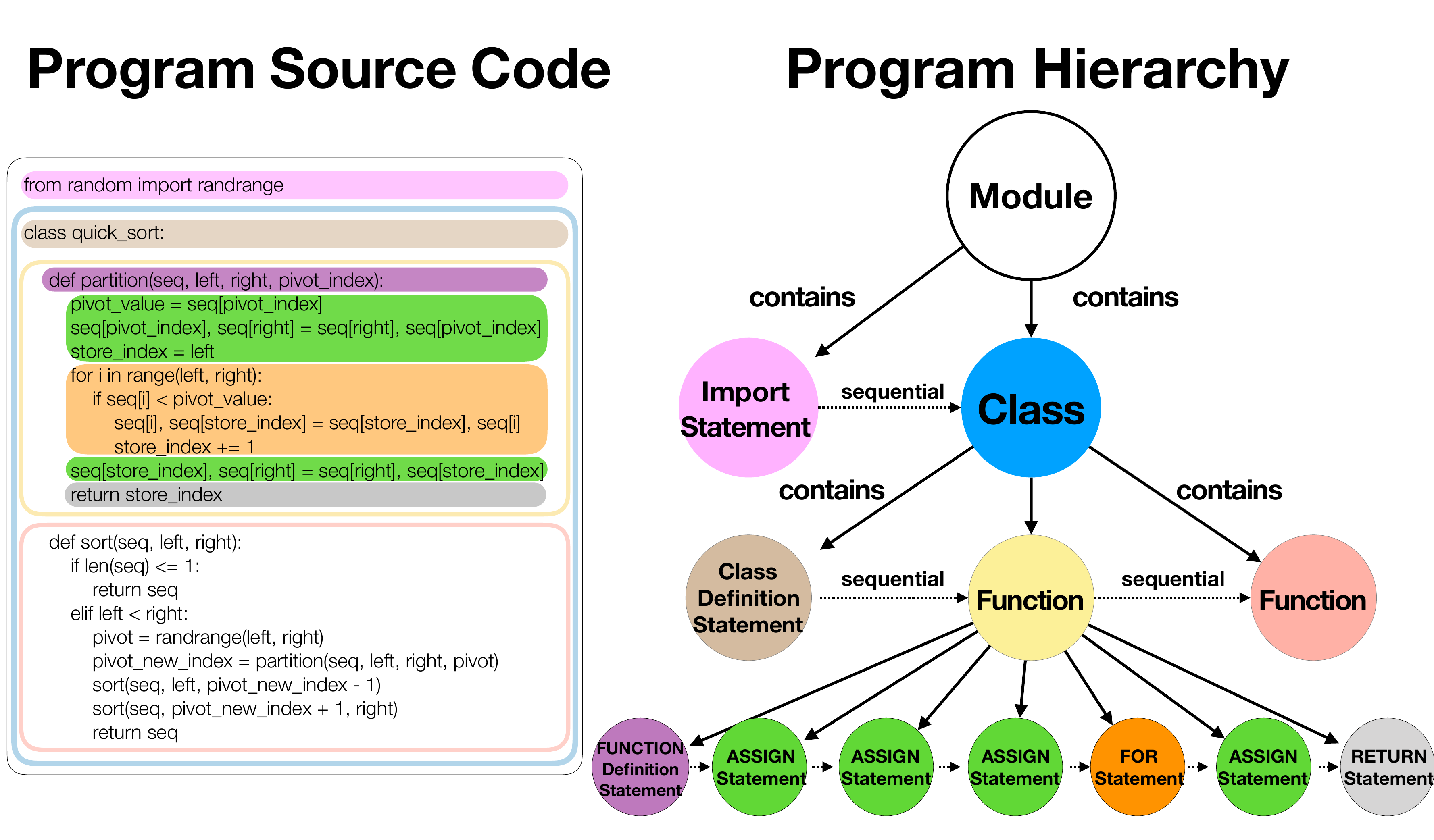}
	\vspace{-10pt}
	\caption{An Example of Program Abstract Syntax Tree.}
	\label{fig:example}
	\vspace{-10pt}
\end{figure}

For instance, as shown in Figure~\ref{fig:example}, given the input program on the left, we can represent its corresponding \textit{program AST} on the right, where the top \textit{program component} is \textit{module}. The \textit{program module} covers the one \textit{import} statement and one \textit{class} component, which further involves two \textit{function} components. The first function componennt \textit{contains} multiple statements with \textit{sequential} relationships, and each statement further \textit{contains} multiple \textit{sequential} expressions, i.e., sequences of tokens. Different from natural language, the program code is well-structured, and each token also has a corresponding concept denoting its type, e.g., \textit{key words} vs \textit{variables} vs \textit{operators}, which can be precisely extracted with the corresponding programming language interpreter/parser.

\subsection{Problem Formulation}

The {\problem} problem studied in this paper actually covers three sub-problems simultaneously, each of which describes a special case of the ``problem synthesis'' problem. Formally, these three sub-problems covered in the {\problem} problem are illustrated as follows:
\begin{itemize}
\item \textit{Program Generation}: Given the \textit{program intention} of the top program module component in the \textit{program AST}, the \textit{program generation} problem aims at generating the program source code that can implement the specified intentions.

\item \textit{Program Interpretation}: With the complete program source code or merely a fragment, the \textit{program interpretation} problem aims at inferring the potential intention of the program, i.e., interpreting the physical functions of the program code.

\item \textit{Program Completion}: Given a fragment of the program code, which can be either a function or merely several statements of the code, the \textit{program completion} problem aims at completing the missing components of the program. 

\end{itemize}

\section{Proposed Methods}\label{sec:method}

In this section, we will introduce the {\our} framework to solve the {\problem} problem (including all these three aforementioned sub-problems). Framework {\our} involves several crucial steps: (1) program parsing, (2) hierarchical sequential statement encoding with {\unit}, and (3) framework learning. In the following part of this section, we will introduce these three steps in great detail.


\subsection{Program Parsing}\label{subsec:parsinng}

Different from natural languages, the program written in programing languages is highly structured. Instead of handling the code characters by characters, we propose to translate the program code into program ASTs in this paper, which will be taken as the input for modeling to be introduced in the next subsection. For instance, given a program statement ``pivot\_value  =  seq[pivot\_index]'', it assigns an entry (with index ``pivot\_index'') from list ``seq'' to a variable ``pivot\_value'', where ``='' and ``[]'' are the operators, and ``pivot\_value'', ``seq'', ``pivot\_index'' denote the assignment target, source list, and index variables respectively. For many programming languages, like Python, the space among the tokens has no impact on the program functions. For instance, the program statement ``pivot\_value=seq[pivot\_index]'' (with no space between the tokens) will work exactly as ``pivot\_value = seq [ pivot\_index ]'' (with tokens well separated by the space). However, such a characteristic will create lots of challenges for partitioning the program line into unit tokens. Traditional text mining and natural language processing techniques will either partition the code line into a sequence of characters, i.e., `p', `i', `v', `o', $\cdots$, `x', `]', or separate the string by certain characters among them. Neither of these two partition methods will work well for programs, and they will also create lots of problems for modeling the program code and understanding the program intention.

In addition, in most of the cases, program operators will be deeply buried in the variables. For instance, the expression ``seq[pivot\_index]'' actually represents an entry in a list, where ``[]'' is an operator. Without differentiating `[' and `]' from the remaining characters, it is highly likely that we will treat ``seq[pivot\_index]'' merely as a new variable name and fail to process the code correctly. In this paper, to resolve such a problem, we propose to parse the program code lines into a program AST instead.

\begin{figure}
\vspace{-30pt}
	\centering
	\includegraphics[width=0.45\textwidth]{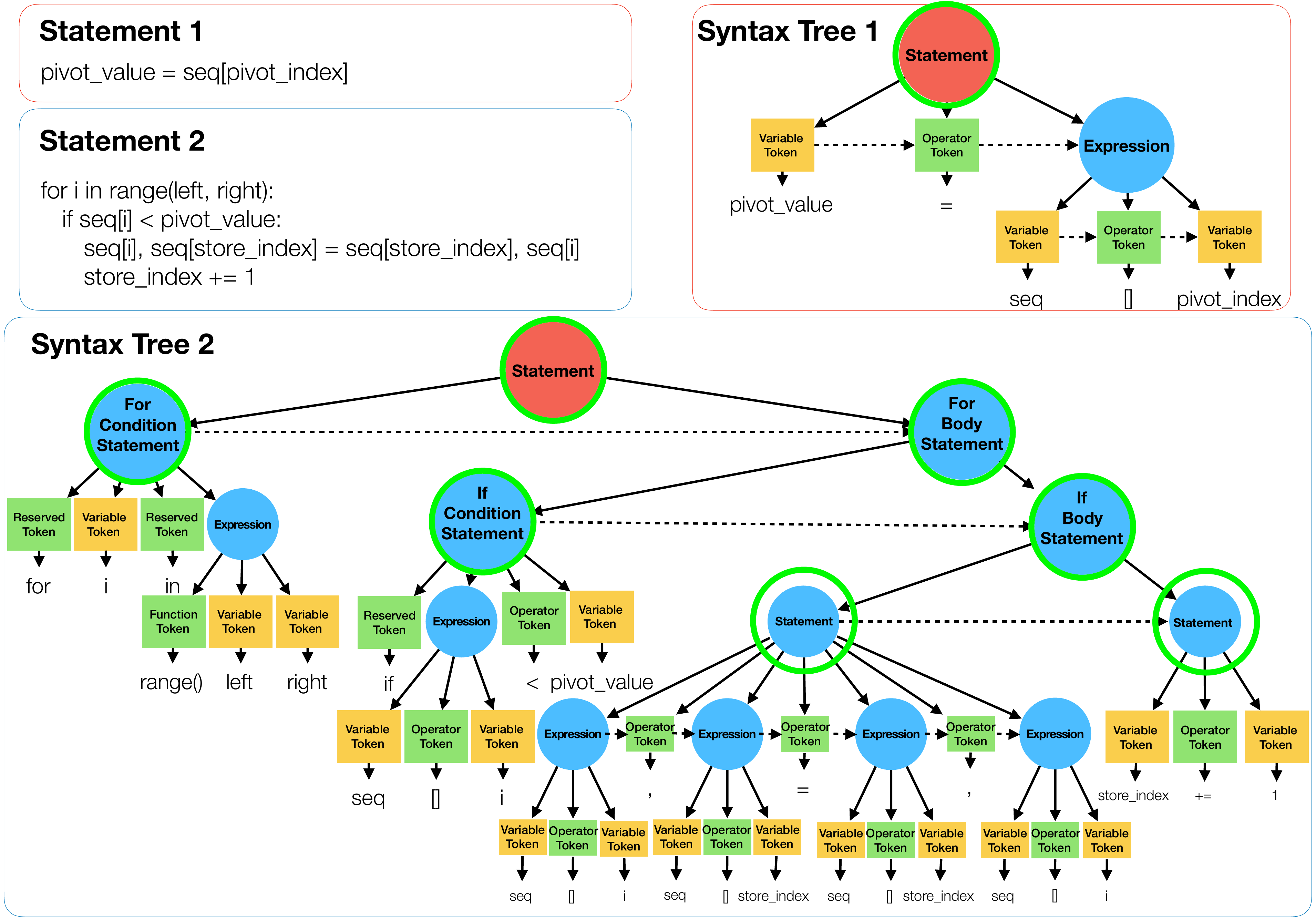}
	\vspace{-10pt}
	\caption{An Example of Program Abstract Syntax Tree.}
	\label{fig:syntax_tree}
	\vspace{-15pt}
\end{figure}

For instance, in Figure~\ref{fig:syntax_tree}, we show two examples of \textit{program ASTs} corresponding to two input program statements. The first statement involves the assignment of value ``seq[pivot\_index]'' to variable ``pivot\_value''. In its AST, we have ``pivot\_value'', ``seq'' and ``pivot\_index'' as the variable tokens, and ``='' and ``[]'' as the operator tokens. Furthermore, ``seq'', ``[]'' and ``pivot\_index'' together will compose an expression in the syntax tree. For the nodes in the same level, i.e., the siblings, we will add sequential links connecting them, which are denoted by the dashed links as shown in Figure~\ref{fig:syntax_tree}. 

The second example shown in Figure~\ref{fig:syntax_tree} is more complicated, it is a ``FOR''-statement. According to the provided syntax tree shown in the figure, this statement contains the ``FOR-Condition''-statement and ``FOR-Body''-statement as the child nodes of the root. For the ``FOR-Condition''-statement, it starts with a reserved keyword token ``for'', followed by variable token and another reserved keyword token ``in'' respectively, and ends with an expression ``range(left, right)'' (involving function call token ``range()'' as well as variable tokens ``left'' and ``right''). Furthermore, in the ``FOR-Body''-statement, it contains an ``IF''-statement, involving both the ``IF-Condition''-statement and ``IF-Body''-statement respectively.


For long programs, their ASTs will be in an extremely deep structure, which may cause many computational problems in model learning. In this paper, we will allow {\our} to truncate ASTs to shrink the tree depth. For instance, if we use \textit{statement} as the smallest basic syntax type in the AST leaf nodes, then the ASTs of program statements 1 and 2 in Figure~\ref{fig:syntax_tree} will be of a much simpler structure, whose involved nodes are marked in green circles in Figure~\ref{fig:syntax_tree}. There exist some open-source tools which can generate the syntax tree of Python code automatically, e.g., the Python AST package\footnote{https://docs.python.org/2/library/ast.html}. With these tools, instead of modeling the program raw textual code, we can translate the program into its AST, and the following learning steps will be all based on the obtained ASTs by default.


\subsection{Hierarchical Sequential Unit (HSU)}

\begin{figure}
\vspace{-30pt}
	\centering
	\includegraphics[width=0.45\textwidth]{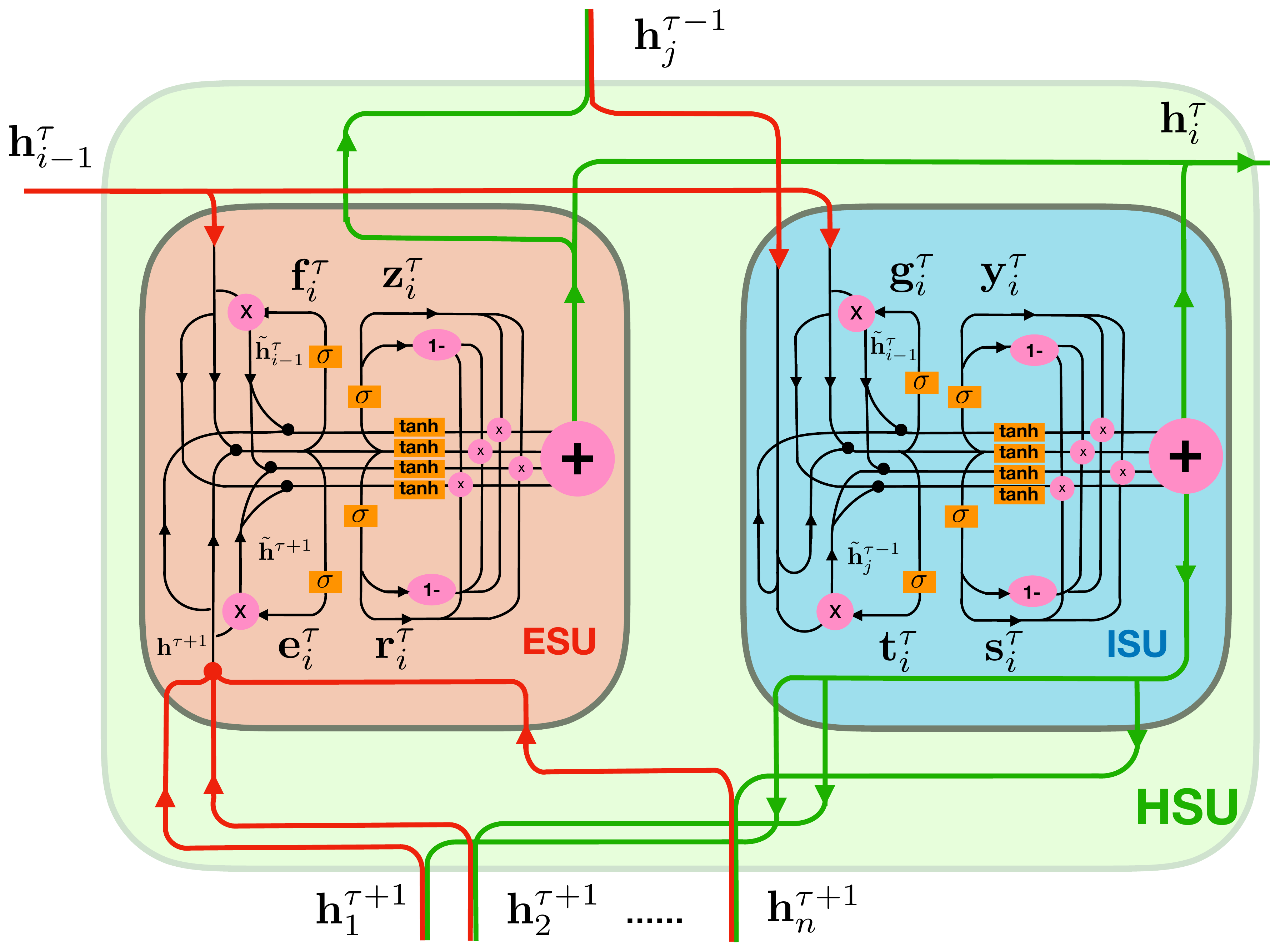}
	\vspace{-10pt}
	\caption{The Hierarchical Sequential Unit (HSU) Model}
	\label{fig:unit_model}
	\vspace{-10pt}
\end{figure}

As shown in the constructed ASTs, among the nodes in the tree structured diagram, there exist two different relationship types: \textit{hierarchical relationship} between the father nodes and children nodes at different levels, and \textit{sequential relationship} between sibling nodes at the same levels. To effectively model the contents of the nodes as well as the hierarchical-sequential relationships among the nodes, in this section, we will introduce a novel unit model, namely HSU (Hierarchical Sequential Unit). {\unit} will be used as the basic structure for constructing the {\our} model (to be illustrated in the next subsection), which involves two sub-units, ESU (Evolutional Sequential Unit) and ISU (Inherited Sequential Unit), for handling the program generation and interpretation tasks respectively. The general structure of the HSU is provided in Figure~\ref{fig:unit_model}, where the arrows denote the information flow directions, black/red dots represent the concatenation operations of vectors, $\sigma$ and $\tanh$ denote the sigmoid and hyperbolic tangent functions respectively, and icons $\otimes$, $\oplus$ represent the entry-wise vector product and sum operators.

\subsubsection{Evolutional Sequential Unit}

In Figure~\ref{fig:unit_model}, the component on the left is an ESU, which accepts the input from the children nodes, i.e., $\mb{h}^{\tau+1}_{1}, \mb{h}^{\tau+1}_{2}, \cdots, \mb{h}^{\tau+1}_{n}$ and the left sibling node, i.e., $\mb{h}^{\tau}_{i-1}$. For the input from sibling node, ESU adopts a ``forget gate'', which may choose one part of $\mb{h}^{\tau}_{i-1}$ to update. In programs, the scope of variables can be different between statements, which may be updated as the code runs into a new statement. Formally, we can represent the ``forget gate'' together with the updated left-sibling node state as
\begin{align*}
\tilde{\mb{h}}^\tau_{i-1} = \mb{f}^\tau_i \otimes \mb{h}^\tau_{i-1}, \mbox{ where  }\mb{f}^\tau_i = \sigma \left( \mb{W}_f \left[{\mb{h}^\tau_{i-1}}, {\mb{h}^{\tau +1}} \right]^\top \right).
\end{align*}
Here, $\mb{h}^{\tau +1} = \left[\mb{h}^{\tau+1}_{1}, \mb{h}^{\tau+1}_{2}, \cdots, \mb{h}^{\tau+1}_{n} \right]$ denotes the concatenated input state vector from the children nodes and matrix $\mb{W}_f$ represents the variables of the ``forget gate'' in ESU.

Meanwhile, for the inputs from the children nodes, ESU introduces a gate, namely the ``evolve gate'', which can evolve the children input states to the upper level. Here, the term ``evolve'' models the changes from the lower-level program expression to higher-level program statement, which is effective to represent the changes in the scope of variables and other program context information across levels in program ASTs. Formally, we can represent the ``evolve gate'' as well as the updated children node state vector as
\begin{align*}
\tilde{\mb{h}}^{\tau+1} = \mb{e}^{\tau}_i \otimes \mb{h}^{\tau+1}, \mbox{ where  }\mb{e}^{\tau}_i = \sigma \left( \mb{W}_e \left[{\mb{h}^\tau_{i-1}}, {\mb{h}^{\tau +1}} \right]^\top \right),
\end{align*}
where $\mb{W}_e$ denotes the variable matrix in the ``evolve gate'' in ESU.

ESU computes the output with the original inputs from sibling and children nodes, i.e., $\mb{h}^{\tau}_{i-1}$, $\mb{h}^{\tau +1}$, as well as the updated sibling-node state vector $\tilde{\mb{h}}^\tau_{i-1}$ and the evolved child-node state vector $\tilde{\mb{h}}^{\tau+1}$. ESU allows different combinations of the state vectors, which are controlled by two new selection gates $\mb{z}_i^\tau$ and $\mb{r}^\tau_i$ respectively. Formally, we can represent the final output of ESU as
\begin{align*}
\mb{h}^{\tau}_i &= \mb{z}_i^\tau \otimes \mb{r}^\tau_i \otimes tanh \left(\mb{W}_u [\tilde{\mb{h}}^\tau_{i-1}, \tilde{\mb{h}}^{\tau+1}]^\top \right)\\
&\oplus (\mb{1} \ominus \mb{z}_i^\tau) \otimes \mb{r}^\tau_i \otimes tanh \left(\mb{W}_u [{\mb{h}}^\tau_{i-1}, \tilde{\mb{h}}^{\tau+1}]^\top \right)\\
&\oplus \mb{z}_i^\tau \otimes (\mb{1} \ominus \mb{r}^\tau_i) \otimes tanh \left(\mb{W}_u [\tilde{\mb{h}}^\tau_{i-1}, {\mb{h}}^{\tau+1}]^\top \right)\\
&\oplus (\mb{1} \ominus \mb{z}_i^\tau) \otimes (\mb{1} \ominus \mb{r}^\tau_i) \otimes tanh \left(\mb{W}_u [{\mb{h}}^\tau_{i-1}, {\mb{h}}^{\tau+1}]^\top \right),
\end{align*}
where $\mb{z}_i^\tau = \sigma ( \mb{W}_z [{\mb{h}^\tau_{i-1}}, {\mb{h}^{\tau +1}} ]^\top ), \mb{r}_i^\tau = \sigma ( \mb{W}_r [{\mb{h}^\tau_{i-1}}, {\mb{h}^{\tau +1}} ]^\top )$, and $\mb{1}$ denotes a vector filled with value $1$. Matrices $\mb{W}_u$, $\mb{W}_z$, $\mb{W}_r$ represent the variables involved in the components. Vector $\mb{h}^{\tau}_i$ will be the output to both the right sibling node and the father node in ESU.

\begin{figure*}
\vspace{-30pt}
	\centering
	\includegraphics[width=0.85\textwidth]{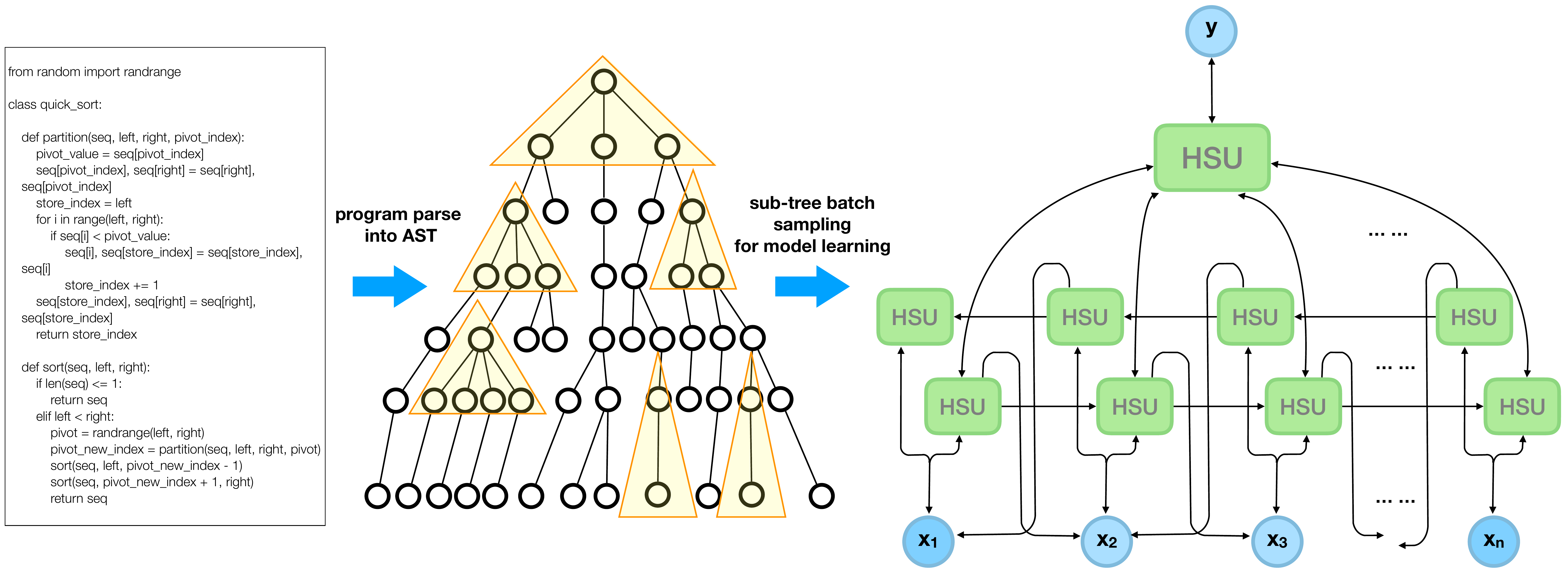}
	\vspace{-10pt}
	\caption{The Architecture of {\our}.}
	\label{fig:architecture}
	\vspace{-10pt}
\end{figure*}

\subsubsection{Inherited Sequential Unit}

The component on the right of Figure~\ref{fig:unit_model} is called the ISU, which accepts input from the left sibling node, i.e., $\mb{h}^{\tau}_{i-1}$, higher-level father node, i.e., $\mb{h}^{\tau-1}_{i}$, and generates the output for the right sibling node and children nodes at the lower level. Similar to ESU, there also exists a ``forget gate'' in ISU for updating some information from the sibling state input. Slightly different from ESU, the ``forget gate'' in ISU is controlled by the states of sibling and father nodes, which together with the updated input from the left-sibling node can be represented as follows:
\begin{align*}
\tilde{\mb{h}}^\tau_{i-1} = \mb{g}^\tau_i \otimes \mb{h}^\tau_{i-1}, \mbox{ where  }\mb{g}^\tau_i = \sigma \left( \mb{W}_g\left[{\mb{h}^\tau_{i-1}}, {\mb{h}^{\tau -1}_j} \right]^\top \right).
\end{align*}
Here, $\mb{W}_g$ is the variable of the ``forget gate'' in ISU.

Another significant difference between ISU and ESU is, for inheriting and updating the program context from the father node, e.g., the scopes of variables and other program information, ISU has an ``inherit gate'' for changing the input states of the father node. Formally, we can represent the ``inherit gate'' together with the updated input from the father node as 
\begin{align*}
\tilde{\mb{h}}^{\tau-1}_j = \mb{t}^{\tau}_i \otimes \mb{h}^{\tau-1}_j, \mbox{ where  }\mb{t}^{\tau}_i = \sigma \left( \mb{W}_t \left[{\mb{h}^\tau_{i-1}}, {\mb{h}^{\tau-1}_j} \right]^\top \right),
\end{align*}
where $\mb{W}_t$ is the variable of the ``inherit gate'' in ISU.

ISU will compute the final output based on the combination of the original input vectors and the updated vectors, which is controlled by the gates $\mb{y}_i^\tau$ and $\mb{s}^\tau_i$ respectively. Formally, we can represent the final output of ISU as
\begin{align*}
\mb{h}^{\tau}_i &= \mb{y}_i^\tau \otimes \mb{s}^\tau_i \otimes tanh \left(\mb{W}_v [\tilde{\mb{h}}^\tau_{i-1}, \tilde{\mb{h}}^{\tau-1}_j]^\top \right)\\
&\oplus (\mb{1} \ominus \mb{y}_i^\tau) \otimes \mb{s}^\tau_i \otimes tanh \left(\mb{W}_v [{\mb{h}}^\tau_{i-1}, \tilde{\mb{h}}^{\tau-1}_j ]^\top \right)\\
&\oplus \mb{y}_i^\tau \otimes (\mb{1} \ominus \mb{s}^\tau_i) \otimes tanh \left(\mb{W}_v [\tilde{\mb{h}}^\tau_{i-1}, {\mb{h}}^{\tau-1}_j ]^\top \right)\\
&\oplus (\mb{1} \ominus \mb{y}_i^\tau) \otimes (\mb{1} \ominus \mb{s}^\tau_i) \otimes tanh \left(\mb{W}_v [{\mb{h}}^\tau_{i-1}, {\mb{h}}^{\tau-1}_j ]^\top \right),
\end{align*}
where gates $\mb{y}_i^\tau = \sigma ( \mb{W}_y [{\mb{h}^\tau_{i-1}}, {\mb{h}^{\tau -1}} ]^\top ), \mb{s}_i^\tau = \sigma ( \mb{W}_s [{\mb{h}^\tau_{i-1}}, {\mb{h}^{\tau -1}} ]^\top )$ and matrices $\mb{W}_y$, $\mb{W}_s$, $\mb{W}_v$ denote the variables of ISU. Vector $\mb{h}^{\tau}_i$ will be the output to both the right sibling node as well as all the children nodes.

In sum, the ESU and ISU components covered in the HSU unit model have a lot in common, as they (1) both have the forget gate, (2) both have the evolve/inherit gate, and (3) both combine the original states and updated states to generate the output. There also exist many difference between ESU and ISU. Besides the input/output among the sibling nodes, ISU also accepts input from higher-level father nodes to generate output to the children nodes; while ESU accepts input from the children nodes instead and generate output to the father node. The evolve/inherit gates in ESU and ISU effectively adapt the program context changes between different levels but in different directions. In the training and testing stages of {\our}, ESU and ISU will be mainly used as the unit structure for program interpretation and program generation to be introduced as follows.


\subsection{Framework Learning}

With the HSU introduced before, we can represent the architecture of {\our} in Figure~\ref{fig:architecture}, which is also in a tree structured diagram. Based on the ASTs parsed from the input program source code, a set of sub-trees will be sampled for training {\our}. For the $n$ children HSU nodes at the lower level, they are fed with their raw encoding features and sibling node states as the inputs. Here, $n=d_{max}$ denotes the maximum node degree in the program AST, and dummy padding will be used for the sub-trees with less than $n$ children nodes. Among these $n$ children nodes, the data flow is bi-directed, which can effectively model the sequential patterns in ASTs in both directions. Furthermore, the outputs of the children HSU nodes will be all fed to a father node at the higher level, which accepts no sibling node input. The output of the father HSU node will effectively recover its content. Besides the bottom-up mode, {\our} can also work well in a top-down mode, where the input of father HSU node will generate the contents of children HSU nodes. In this part, we will introduce the {\our} model in great detail to illustrate how to train the model with program ASTs.

\subsubsection{Token Raw Encoding} 

As introduced before, in the program ASTs, the nodes denote the program components, which contain program syntax types, token contents and program intentions (optional). Based on the parsing results obtained from the program, we can obtain the syntax type set and the set of concrete keyword, variable, operator and other tokens used in the program, which will be represented as sets $\mathcal{C}$ and $\mathcal{T}$ respectively. Formally, for each node $v_i \in \mathcal{V}$ in the program ASTs, its representation can be represented as a vector $\mb{x}_{i} = [\mb{x}_{i}^c, \mb{x}_{i}^t] \in \{0, 1\}^{|\mathcal{C}| + k \cdot |\mathcal{T}|}$, where $\mb{x}_{i}^c \in \{0, 1\}^{|\mathcal{C}|}$ and $\mb{x}_{i}^t \in \{0, 1\}^{k \cdot |\mathcal{T}|}$ represent the one-hot feature vector of syntax types and tokens respectively and $k$ represents the maximal number of tokens contained in the tree nodes. For the tree nodes with less than $k$ tokens, dummy padding will be adopted.

\subsubsection{Program Generation: Top-Down Training of {\our}}

Based on the input raw feature vector $\mb{y}$ from the father node in {\our} as illustrated in Figure~\ref{fig:architecture}, we can denote its output result of the father node via the ISU model as 
$$\mb{h}^{\tau} = \mbox{ISU}(\mb{y}, \mb{null}; \mb{W}_{I}),$$
where $\mb{null}$ denotes a dummy padding vector and $\mb{W}_I$ covers all the variables involved in the ISU model introduced before.

By feeding $\mb{h}^{\tau}$ as the input to the children nodes at the lower level, model {\our} will generate the output representations of the children nodes. We can denote the state and output vectors of the $i_{th}$ child node as 
$$\begin{cases}
\mb{h}_i^{\tau+1} &= \mbox{ISU} (\mb{h}^{\tau}, \mb{h}^{\tau+1}_{i-1}; \mb{W}_{I} ),\\
\hat{\mb{x}}_i &= \mbox{softmax}(\mb{W}_{down} \mb{h}_i^{\tau+1} + \mb{b}_{down}),
\end{cases} $$
where $\mb{h}^{\tau+1}_{i-1}$ denotes the input from the left sibling node, $\mbox{softmax}(\cdot)$ represents the softmax function and $\mb{h}^{\tau+1}_{0} = \mb{null}$ for the first child node without left sibling. $\mb{W}_{down}$ and $\mb{b}_{down}$ are the variables involved to project node state to the output.

Compared with the ground-truth representation of the children nodes in the sampled sub-tree, i.e., $\{ \mb{x}_i \}_{i=1}^{d_{max}}$, the loss introduced by the ISU model on the sub-tree can be represented as
\begin{equation}
\mathcal{L}_{\mbox{\textsc{isu}}} = \sum_{i=1}^{ d_{max} } \sum_{j = 1}^{|\mathcal{C}| + k\cdot |\mathcal{T}|} - \mb{x}_i [j] \log \hat{\mb{x}}_i [j],
\end{equation}
which is defined based on the cross-entropy loss function, and index $j$ enumerates all the syntax types and tokens involved in the node representation vector.

\subsubsection{Program Interpretation: Bottom-Up Training of {\our}}

On the other hand, besides the top-down direction, we will also train the {\our} model in a bottom-up manner. Based on the input for the children nodes, we can generate the contents of the father node as well. Formally, we can represent the input vectors for the children nodes as $\{ \mb{x}_i \}_{i = 1}^{d_{max}}$. By feeding these vectors to the children nodes, we can represent the output vector from the $i_{th}$ child node as vector $\mb{h}_i^{\tau+1}$:
$$ \mb{h}_i^{\tau+1} = \mbox{ESU}(\mb{x}_i, \mb{h}_{i-1}^{\tau+1}; \mb{W}_{E}), $$
where $\mb{h}_{0}^{\tau+1} = \mb{null}$ for the first children node, and $\mb{W}_E$ represents the variables involved in the ESU model.

Furthermore, based on the children node representations, we will be able to represent the state vector and output vector of the father node as
$$
\begin{cases}
\mb{h}^\tau &= \mbox{ESU}( \mb{h}^{\tau+1}, \mb{null}; \mb{W}_{E} ),\\
\hat{\mb{y}} &= \mbox{softmax}( \mb{W}_{up} \mb{h}^\tau + \mb{b}_{up} ),
\end{cases}
$$
where vector $\mb{h}^{\tau+1} = [\mb{h}^{\tau+1}_1, \mb{h}^{\tau+1}_2, \cdots, \mb{h}^{\tau+1}_n]^\top$ contains all the children node states. $\mb{W}_{up}$ and $\mb{b}_{up}$ are the variables used to project the father node state to the its output.

The introduced loss based on the input sub-tree by comparing $\hat{\mb{y}}$ with the ground-truth vector $\mb{y}$ can be represented as
\begin{equation}
\mathcal{L}_{\mbox{\textsc{esu}}} = \sum_{j = 1}^{|\mathcal{C}| + k\cdot |\mathcal{T}|} - \mb{y} [j] \log \hat{\mb{y}} [j].
\end{equation}

\subsubsection{Program Completion: Sequential Training of {\our} }

In the case when only a fragment of the program is provided for feeding the child nodes, the training process for the ESU will encounter great challenges, since the incomplete input will mislead the model to generate a wrong output. This happens very often, since missing any line of the program code will introduce an incomplete sub-tree structured diagram in the program AST. To resolve such a problem, we propose to generate the complete child node input information based on the program fragments by training the bi-directed HSU structure in {\our}.

As introduced before, based on the input of children nodes $\{ \mb{x}_i \}_{i=1}^{d_{max}}$, we can represent their state vectors as $\{ \mb{h}_i^{\tau+1} \}_{i=1}^{d_{max}}$. In the bi-directed HSU, based on the state vectors of the $i_{th}$ child node, we can represent the inferred output vectors for the tokens on the left and on the right as vectors $\hat{\mb{x}}_{i, l}$ and $\hat{\mb{x}}_{i, r}$ respectively:
$$
\begin{cases}
\hat{\mb{x}}_{i, l} = \mbox{softmax}( \mb{W}_{left} \mb{h}_i^{\tau+1} + \mb{b}_{left} ),\\
\hat{\mb{x}}_{i, r} = \mbox{softmax}( \mb{W}_{right} \mb{h}_i^{\tau+1} + \mb{b}_{right} ),\\
\end{cases}
$$
where $\mb{W}_{left}$, $\mb{b}_{left}$, and $\mb{W}_{right}$, $\mb{b}_{right}$ are the variables used to project the states to the output in the left and right HSUs respectively. Compared with the ground truth, we can represent the loss introduced in generating children node tokens as:
\begin{align}
\mathcal{L}_{\mbox{\textsc{hsu}}} &= \sum_{i = 2}^{d_{max}} \sum_{j = 1}^{(|\mathcal{C}| + k\cdot |\mathcal{T}| )}  -{\mb{x}}_{i}[j] \log \hat{\mb{x}}_{i-1, r}[j] \\
&+ \sum_{i = 1}^{(d_{max}-1)} \sum_{j = 1}^{(|\mathcal{C}| + k\cdot |\mathcal{T}|)}  -{\mb{x}}_{i}[j] \log \hat{\mb{x}}_{i+1, l}[j].
\end{align}

\subsubsection{Joint Optimization Objective Function}

Based on the above descriptions, we can represent the joint optimization function of model {\our} as 
\begin{equation}
\min_{\mb{W}_I, \mb{W}_E, \mb{W}_P} \mathcal{L}_{\mbox{\textsc{isu}}} + \alpha \cdot \mathcal{L}_{\mbox{\textsc{esu}}} + \beta \cdot \mathcal{L}_{\mbox{\textsc{hsu}}},
\end{equation}
where $\mb{W}_P$ covers all the variables adopted to project the state vector to the output space introduced above, and $\alpha$, $\beta$ denote the weights of the last two loss terms (in the experiments $\alpha$ and $\beta$ are both assigned with value $1.0$).

Formally, to solve the above objective function, the learning process of {\our} can be done based on the sub-tree structures (involving one parent node and all its child nodes) sampled from the program AST. To optimize the above loss function, we utilize Stochastic Gradient Descent (SGD) as the optimization algorithm. To be more specific, the training process involves multiple epochs. In each epoch, the training data is shuffled and a minibatch of the instances are sampled to update the parameters with SGD. In addition, for each sampled sub-tree, we will feed the {\our} model to minimize the loss terms $\mathcal{L}_{\mbox{\textsc{isu}}}$, $\mathcal{L}_{\mbox{\textsc{esu}}}$ and $\mathcal{L}_{\mbox{\textsc{hsu}}}$ iteratively for parameter learning. Such a process continues until convergence.

\section{Experiments}\label{sec:experiment}

To test the effectiveness of the proposed unit model {\unit} and the learning framework {\our}, we have conducted extensive experiments on a real-world program-comment dataset, and compared {\our} with several existing text generation methods. In the following part of this section, we will first introduce the experimental settings, including dataset descriptions, detailed experiment setup, comparison methods and evaluation metrics. After that, the experimental results and case studies will be provided and analyzed.

\begin{table*}[t]
\vspace{-30pt}
\caption{Next Line Program Code Inference.} 
\label{tab:generation}
\vspace{-10pt}
\scriptsize
\centering
\setlength{\tabcolsep}{3pt}
{ \begin{tabular}{rccccccccccc}
\toprule
&\multicolumn{11}{c}{Evaluation Metrics}\\
\cmidrule{2-12}
methods	&Accuracy	&Mi-Precision	&Mi-Recall	&Mi-F1 &Ma-Precision	&Ma-Recall	&Ma-F1 &W-Precision	&W-Recall	&W-F1  &Train-Iteration \\ 
\midrule

{\our}	&\textbf{0.949}$\pm$\textbf{0.149}	&\textbf{0.949}$\pm$\textbf{0.149}	&\textbf{0.949}$\pm$\textbf{0.149}	&\textbf{0.949}$\pm$\textbf{0.149}	&\textbf{0.932}$\pm$\textbf{0.186}	&\textbf{0.93}$\pm$\textbf{0.188}	&\textbf{0.93}$\pm$\textbf{0.189}	&\textbf{0.954}$\pm$\textbf{0.143}	&\textbf{0.949}$\pm$\textbf{0.149}	&\textbf{0.95}$\pm$\textbf{0.148}	 &\textbf{43,000}\\

\cmidrule{2-12}

{\astbilstm}	&0.86$\pm$0.195	&0.86$\pm$0.195	&0.86$\pm$0.195	&0.86$\pm$0.195	&0.803$\pm$0.253	&0.819$\pm$0.237	&0.806$\pm$0.249	&0.847$\pm$0.216	&0.86$\pm$0.195	&0.847$\pm$0.21	&86,000\\
{\astbigru}	&0.858$\pm$0.191	&0.858$\pm$0.191	&0.858$\pm$0.191	&0.858$\pm$0.191	&0.798$\pm$0.249	&0.817$\pm$0.233	&0.802$\pm$0.244	&0.843$\pm$0.213	&0.858$\pm$0.191	&0.844$\pm$0.207	&86,000\\
{\astbirnn}	&0.804$\pm$0.212	&0.804$\pm$0.212	&0.804$\pm$0.212	&0.804$\pm$0.212	&0.734$\pm$0.26	&0.741$\pm$0.254	&0.732$\pm$0.259	&0.807$\pm$0.222	&0.804$\pm$0.212	&0.798$\pm$0.22	&86,000\\

\cmidrule{2-12}

{\autoencoder}	&0.326$\pm$0.175	&0.326$\pm$0.175	&0.326$\pm$0.175	&0.326$\pm$0.175	&0.175$\pm$0.14	&0.222$\pm$0.149	&0.186$\pm$0.139	&0.276$\pm$0.187	&0.326$\pm$0.175	&0.284$\pm$0.174	&4,300,000\\

\cmidrule{2-12}
{\bilstm}	&0.774$\pm$0.07	&0.774$\pm$0.07	&0.774$\pm$0.07	&0.774$\pm$0.07	&0.753$\pm$0.126	&0.76$\pm$0.115	&0.756$\pm$0.122	&0.767$\pm$0.087	&0.774$\pm$0.07	&0.769$\pm$0.081	&430,000	\\
{\bigru}	&0.774$\pm$0.07	&0.774$\pm$0.07	&0.774$\pm$0.07	&0.774$\pm$0.07	&0.753$\pm$0.126	&0.76$\pm$0.115	&0.756$\pm$0.122	&0.767$\pm$0.087	&0.774$\pm$0.07	&0.769$\pm$0.081	&430,000	\\
{\birnn}	&0.749$\pm$0.08	&0.749$\pm$0.08	&0.749$\pm$0.08	&0.749$\pm$0.08	&0.707$\pm$0.148	&0.71$\pm$0.148	&0.708$\pm$0.148	&0.746$\pm$0.082	&0.749$\pm$0.08	&0.747$\pm$0.081	&430,000 \\
\bottomrule
\end{tabular}}
\vspace{-10pt}
\end{table*}

\begin{table}[t]
\caption{Program Generation from Comments.}
\label{tab:generation_complete}
\vspace{-10pt}
\footnotesize
\centering
\setlength{\tabcolsep}{3pt}
{ \begin{tabular}{rc}
\toprule
methods	&Accuracy	 \\ 
\midrule
{\our}		&35/43	\\

\cmidrule{2-2}
{\astbilstm}	&13/43	\\
{\astbigru}		&13/43	\\
{\astbirnn}		&11/43	\\

\cmidrule{2-2}
{\autoencoder}	&0/43	\\

\cmidrule{2-2}
{\bilstm}		&2/43	\\
{\bigru}		&1/43	\\
{\birnn}		&2/43	\\

\bottomrule
\end{tabular} }
\vspace{-10pt}
\end{table}

\subsection{Experimental Setting}


\subsubsection{Dataset Description}

In the experiments, we will take the program code written in Python programing language as an example. The program dataset used in the experiments covers the Python implementation code of basic algorithms, like different sort algorithms, search algorithms, hash algorithms and dynamic program algorithms. In the program source code file, besides the source code, there also exist a sequence of comments indicating the functions of the program, which will be used as the program intention in the experiments. The dataset will be released as a benchmark for code generation soon.

\subsubsection{Experimental Setup}

In the experiments, instead of modeling the program code characters by characters, we propose to parse the code into ASTs, in which the smallest syntax type is the basic statement in the experiments. From the ASTs, we can sample a set of sub-tree structured diagrams. The contents attached to the sub-tree nodes together with its structure will be fed to learn the {\our} model. Based on the program ASTs, we can denote the maximum children node number as $d_{max}$, and the maximum number of tokens attached to each node as $k$. For the nodes with less than $d_{max}$ nodes or $k$ tokens, a dummy one-hot key feature vector representation will be used for padding. At the same time, for the AST root node, we extract the textual comments of the whole program as its content, which will be represented as a sequence of natural language tokens actually, and can be modeled with the {\unit} model effectively as well. Based on the sampled sub-trees, we propose to train {\our} iteratively as introduced at the end of Section~\ref{sec:method} to learn the variables.

\subsubsection{Comparison Methods}

In the experiments, we will compare {\our} with various existing prediction and generative models, which are listed as follows:

\begin{itemize}
\item \textit{{\our} Model}: The {\our} model proposed in this paper is based on the new {\unit} unit model, which can learn the information in program code based on its ASTs.

\item \textit{\astbilstm}: The {\astbilstm} model is an AST based bi-directional RNN model \cite{SP97} using LSTM \cite{HS97} as the unit cell. {\astbilstm} can capture the sequential patterns in program code textual data in bi-directions simultaneously, which is able to infer the tokens ahead of and after the input.

\item \textit{\astbigru}: The {\astbigru} model is also an AST based bi-directional RNN \cite{SP97} model using GRU \cite{CGCB14} as the unit cell. {\astbigru} can capture the sequential patterns in program code textual data in bi-directions as well.

\item \textit{\astbirnn}: In the experiments, we also compare with the bi-directional RNN model {\astbirnn} \cite{SP97}, which uses the basic neuron cell as the unit cell.

\item \textit{{\autoencoder}}: Via two hidden layers, we propose to use the deep Autoencoder model \cite{VLLBM10} as another baseline method in the experiments. {\autoencoder} can effectively capture the patterns for sequential program components extracted from the ASTs.

\item \textit{\bilstm}: To show the advantages of modeling the program code textual information based on the AST, we also compare the methods with the traditional bi-directional RNN \cite{SP97} models based on the raw program textual information. Model {\bilstm} splits the program code and comment into tokens based on the space among them.

\item \textit{\bigru}: Model {\bigru} using GRU \cite{CGCB14} as the unit model also splits the program code and comment into tokens based on the space among them and infers the following tokens iteratively. 

\item \textit{\birnn}: Model {\birnn} is of the same architecture as {\bilstm} and {\bigru}, but it uses the basic neuron as the unit model in the learning process.


\end{itemize}

\begin{table}[t]
\caption{Program Completion from a Random Input Line.}
\label{tab:completion}
\vspace{-10pt}
\centering
\footnotesize
\setlength{\tabcolsep}{3pt}
{
\begin{tabular}{rc}
\toprule
methods	&Accuracy	 \\ 
\midrule
{\our}		&34/43	\\

\cmidrule{2-2}
{\astbilstm}	&12/43	\\
{\astbigru}		&12/43	\\
{\astbirnn}		&10/43	\\

\cmidrule{2-2}
{\autoencoder}	&0/43	\\

\cmidrule{2-2}
{\bilstm}		&0/43	\\
{\bigru}		&0/43	\\
{\birnn}		&0/43	\\

\bottomrule
\end{tabular}
}\vspace{-10pt}
\end{table}

\begin{table*}[t]
\vspace{-30pt}
\caption{Program Interpretation with Input Source Code.} 
\label{tab:interpretation}
\vspace{-10pt}
\scriptsize
\centering
\setlength{\tabcolsep}{3pt}
{
\begin{tabular}{rccccccccccc}
\toprule
&\multicolumn{11}{c}{Evaluation Metrics}\\
\cmidrule{2-12}
methods	&Accuracy	&Mi-Precision	&Mi-Recall	&Mi-F1 &Ma-Precision	&Ma-Recall	&Ma-F1 &W-Precision	&W-Recall	&W-F1 &Train-Iteration \\ 
\midrule
{\our}	&\textbf{0.978}$\pm$\textbf{0.072}	&\textbf{0.978}$\pm$\textbf{0.072}	&\textbf{0.978}$\pm$\textbf{0.072}	&\textbf{0.978}$\pm$\textbf{0.072}	&\textbf{0.963}$\pm$\textbf{0.107}	&\textbf{0.963}$\pm$\textbf{0.105}	&\textbf{0.963}$\pm$\textbf{0.107}	&\textbf{0.978}$\pm$\textbf{0.072}	&\textbf{0.978}$\pm$\textbf{0.072}	&\textbf{0.977}$\pm$\textbf{0.073}	 &\textbf{43,000}\\
\cmidrule{2-12}
{\astbilstm}	&0.84$\pm$0.198	&0.84$\pm$0.198	&0.84$\pm$0.198	&0.84$\pm$0.198	&0.779$\pm$0.248	&0.793$\pm$0.241	&0.78$\pm$0.248	&0.832$\pm$0.21	&0.84$\pm$0.198	&0.829$\pm$0.208	&86,000\\
{\astbigru}	&0.849$\pm$0.188	&0.849$\pm$0.188	&0.849$\pm$0.188	&0.849$\pm$0.188	&0.787$\pm$0.241	&0.802$\pm$0.232	&0.789$\pm$0.24	&0.84$\pm$0.202	&0.849$\pm$0.188	&0.837$\pm$0.198	&86,000	\\
{\astbirnn}	&0.798$\pm$0.212	&0.798$\pm$0.212	&0.798$\pm$0.212	&0.798$\pm$0.212	&0.728$\pm$0.254	&0.731$\pm$0.254	&0.724$\pm$0.257	&0.805$\pm$0.213	&0.798$\pm$0.212	&0.793$\pm$0.215	&86,000 \\
\cmidrule{2-12}
{\autoencoder}	&0.321$\pm$0.182	&0.321$\pm$0.182	&0.321$\pm$0.182	&0.321$\pm$0.182	&0.174$\pm$0.144	&0.214$\pm$0.151	&0.183$\pm$0.143	&0.279$\pm$0.192	&0.321$\pm$0.182	&0.285$\pm$0.182	&4,300,000\\
\cmidrule{2-12}
{\bilstm}	&0.782$\pm$0.069	&0.782$\pm$0.069	&0.782$\pm$0.069	&0.782$\pm$0.069	&0.769$\pm$0.121	&0.769$\pm$0.121	&0.769$\pm$0.121	&0.782$\pm$0.069	&0.782$\pm$0.069	&0.782$\pm$0.069	&430,000\\
{\bigru}	&0.782$\pm$0.069	&0.782$\pm$0.069	&0.782$\pm$0.069	&0.782$\pm$0.069	&0.769$\pm$0.121	&0.769$\pm$0.121	&0.769$\pm$0.121	&0.782$\pm$0.069	&0.782$\pm$0.069	&0.782$\pm$0.069	&430,000\\
{\birnn}	&0.751$\pm$0.135	&0.751$\pm$0.135	&0.751$\pm$0.135	&0.751$\pm$0.135	&0.731$\pm$0.183	&0.731$\pm$0.183	&0.731$\pm$0.183	&0.751$\pm$0.135	&0.751$\pm$0.135	&0.751$\pm$0.135	&430,000\\
\bottomrule
\end{tabular}
}
\end{table*}


\begin{figure*}[t]
\vspace{-10pt}
\centering
\subfigure[Binary Search Code.]{ \label{fig:search}
    \begin{minipage}[l]{0.45\columnwidth}
      \centering
      \includegraphics[width=1.0\textwidth]{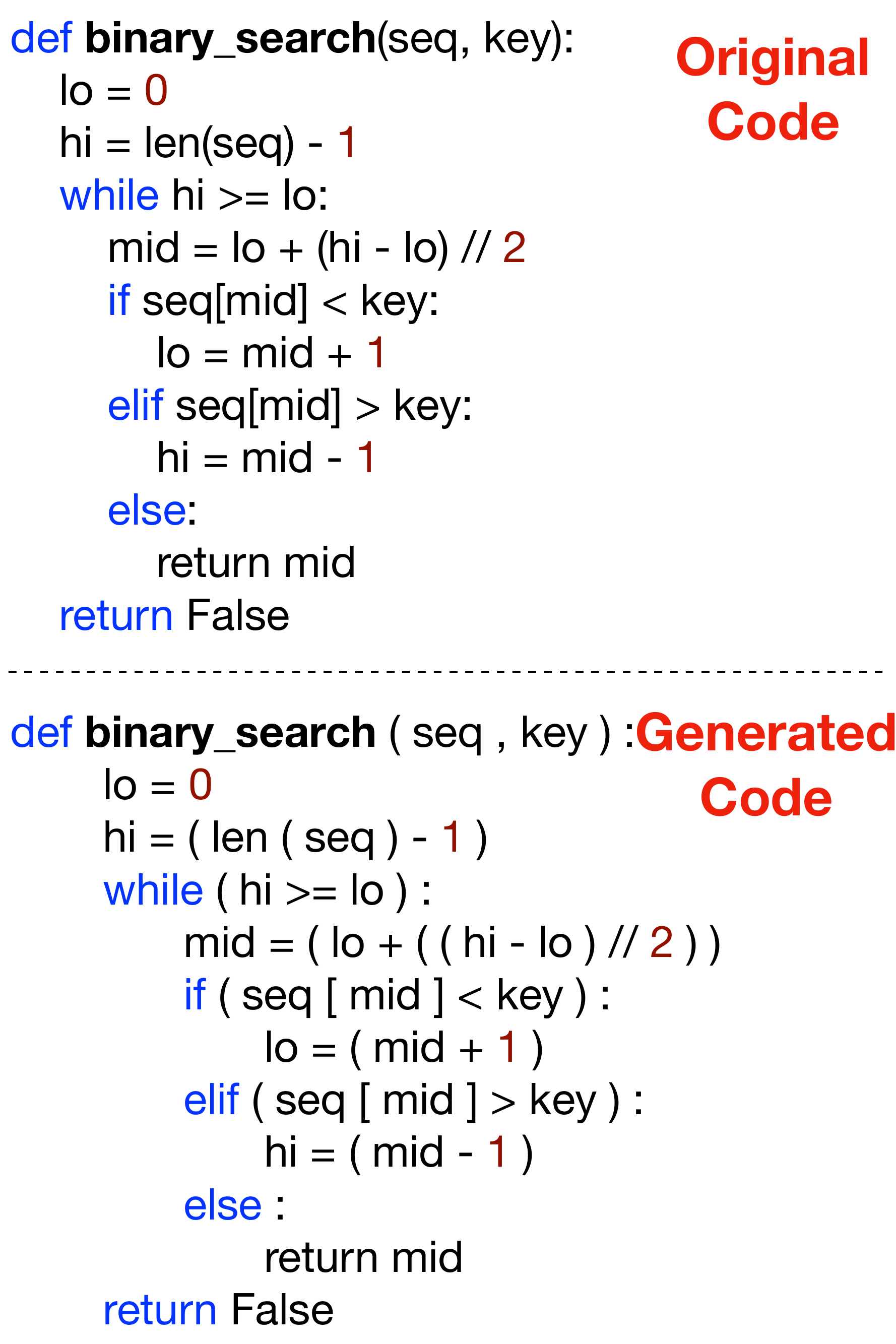}
    \end{minipage}
}
\subfigure[Quick Sort Code.]{\label{fig:sort}
    \begin{minipage}[l]{1.55\columnwidth}
      \centering
      \includegraphics[width=1.0\textwidth]{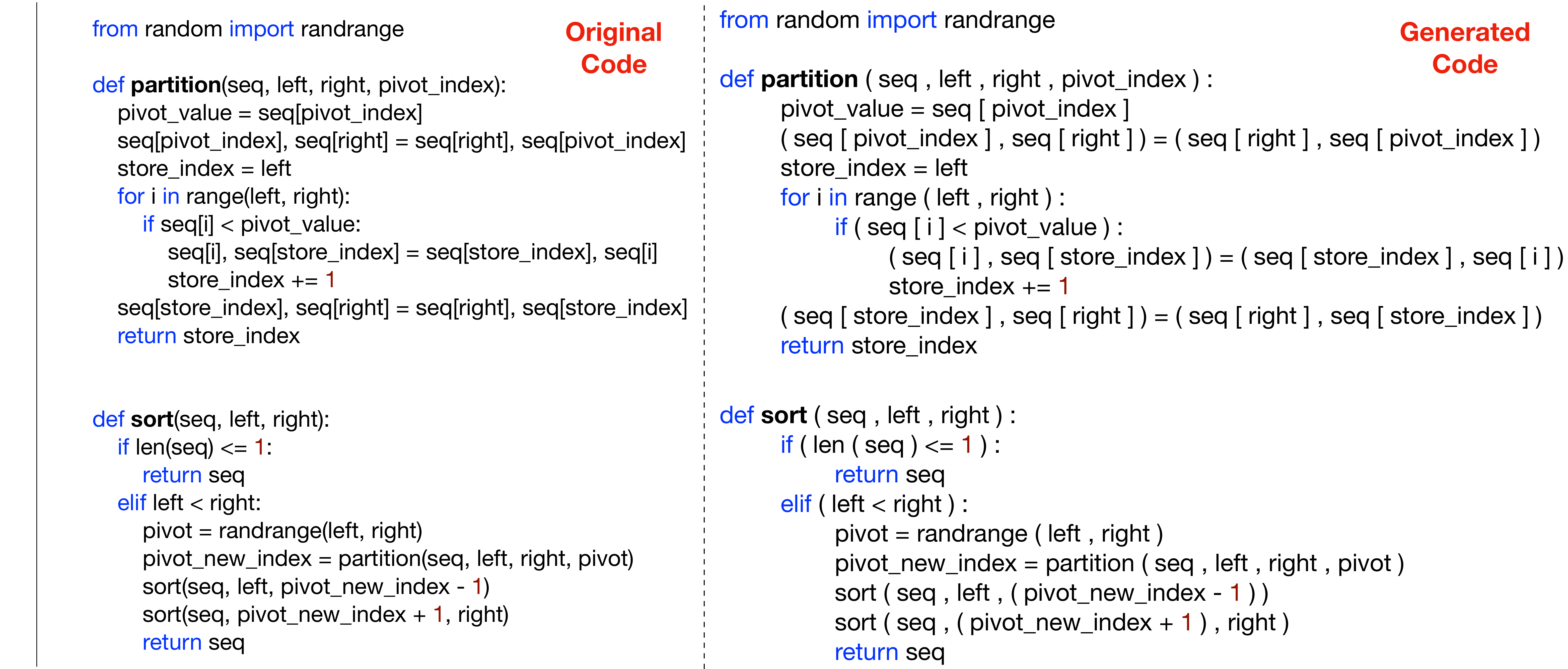}
    \end{minipage}
}
\vspace{-10pt}
\caption{Succeeded Examples of {\our}. Left: Binary Search; Right: Quick Sort.}\label{fig:succeeded}
\vspace{-10pt}
\end{figure*}

\subsubsection{Evaluation Metrics}

We formulate the program code generation (from comments to program code), program code interpretation (from program code to program comments), and program completion problems as a multi-class classification problem, where the inferred tokens can be used as labels. In the experiments, we will use traditional classification evaluation metric Accuracy, Precision, Recall and F1 for measuring the performance of models. Here, the Precision, Recall and F1 metrics cover the micro, macro and weighted versions respectively. In addition, we will also show the number of iterations required in training the models as another evaluation metric. Here, we need to add a remark that, as indicated in page\footnote{http://scikit-learn.org/stable/modules/generated/sklearn.metrics.precision\_recall\_fscore\_support.html}, for the weighted-F1 metric which considers the label imbalance, its value may not be between the corresponding weighted precision and recall in the experimental results.

\subsection{Experimental Result}

\subsubsection{Program Generation from Comments}

In Table~\ref{tab:generation}, we show the performance of different methods in generating the program code (line by line). Based on the input program code line, the models will predict the next line of the program code, and the inferred tokens in the new line are treated as the labels in evaluation.

According to the results in Table~\ref{tab:generation}, model {\our} can achieve much better performance generating the program code line compared with the other baseline methods. For instance, the Accuracy achieved by {\our} is $0.949$, which is almost the triple of the Accuracy achieved by {\autoencoder} and also surpasses {\astbilstm}, {\astbigru} and {\astbirnn} by more than $10\%$ and outperforms {\bilstm}, {\bigru} and {\birnn} by more than $22.6\%$. Similar results can also be observed for the other evaluation metrics. In addition, model {\our} takes far less iterations before convergence based on the training set. As shown in Table~\ref{tab:generation}, the iteration required for {\our} to converge is merely about $43,000$, which is about $\frac{1}{100}$ of the required iterations by {\autoencoder}, $\frac{1}{2}$ of the required iterations by {\astbilstm}, {\astbigru}, {\astbirnn}, and about $\frac{1}{10}$ of the required iterations by {\bilstm}, {\bigru} and {\birnn}. 

In Table~\ref{tab:generation_complete}, we show the results obtained by {\our} in generating the complete program code based on the input program comments. Here, the generation process involves the iterative inferences of the program tokens at the next lines based on the input program comments without any interactions with the outside world. Among the $43$ input program comments, {\our} is able to generate $35$ of them without making any mistakes, which outperform the baseline methods with great advantages. For the AST-BiRNN methods, they can generate $11$-$13$ of the program without any mistakes. For the traditional BiRNN methods, they can only generate 1-2 programs, while {\autoencoder} cannot generate any program at all.

\subsubsection{Program Interpretation based on Code}

In Table~\ref{tab:interpretation}, we provide the experimental results of the comparison methods in generating the program comments based on the program code input. Compared with the program code, the program comments are of a shorter length and have less tokens to be predicted, and the evaluation scores achieved by the baseline methods in Table~\ref{tab:interpretation} are also slightly larger than the scores in Table~\ref{tab:generation}.

Among the baseline methods, {\our} can still achieve much better results than the baseline methods with great advantages. Among all the program code input, {\our} can correctly generate about $0.978$ the program comment tokens. The AST-BiRNN methods can also achieve a very good performance, and they can obtain an average Accuracy around $0.8$, which is better than the traditional BiRNN methods. Method {\autoencoder} performs the worst in the program interpretation task, which can merely achieve an average Accuracy score around $0.321$. 

\begin{figure*}[t]
\vspace{-30pt}
\centering
\subfigure[Sieve of Atkin Code.]{ \label{fig:atkin}
	\begin{minipage}[l]{0.66\columnwidth}
      \centering
      \includegraphics[width=1.0\textwidth]{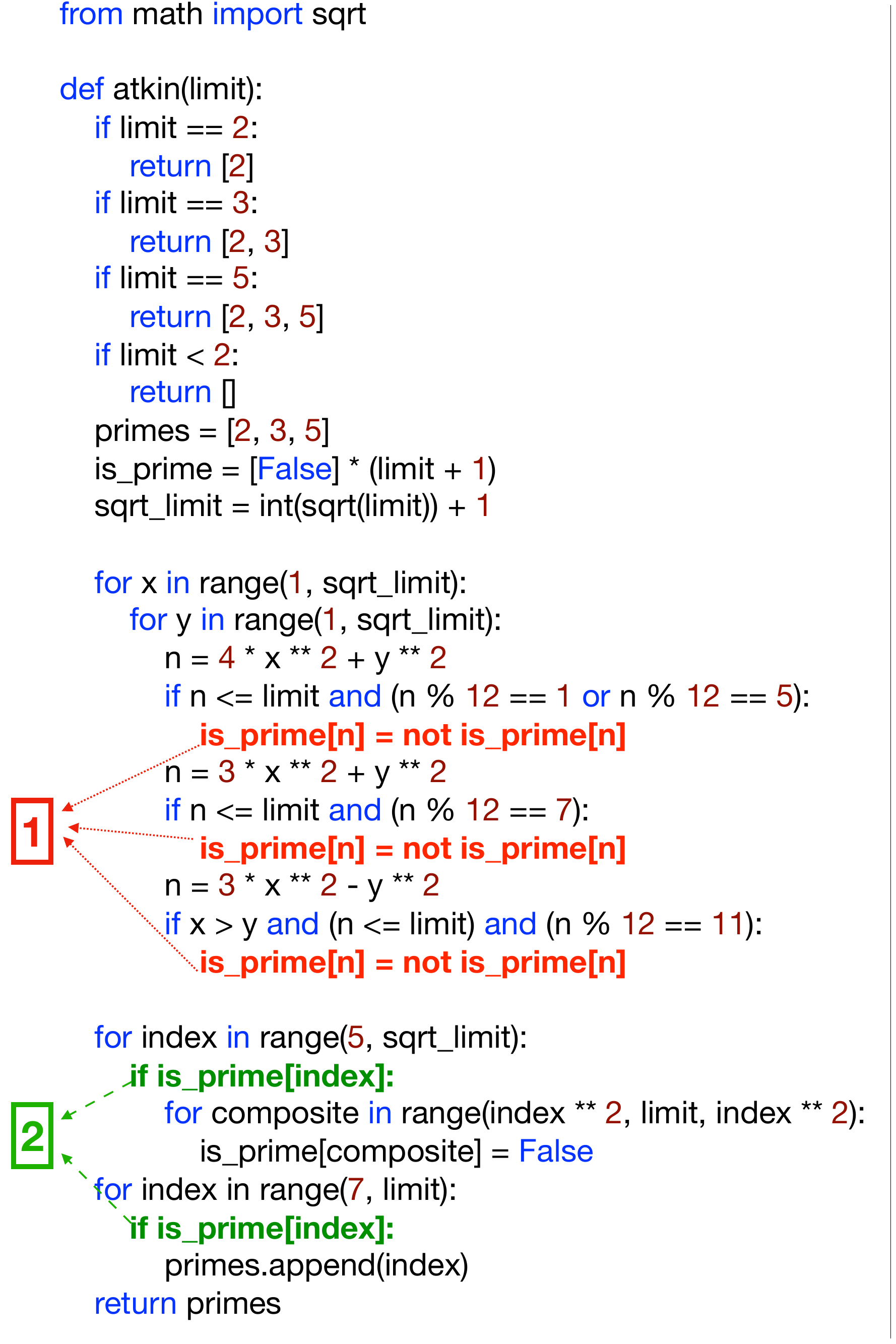}
    \end{minipage}
}
\subfigure[Union Find Class Code.]{ \label{fig:unionfind}
	\begin{minipage}[l]{1.33\columnwidth}
      \centering
      \includegraphics[width=1.0\textwidth]{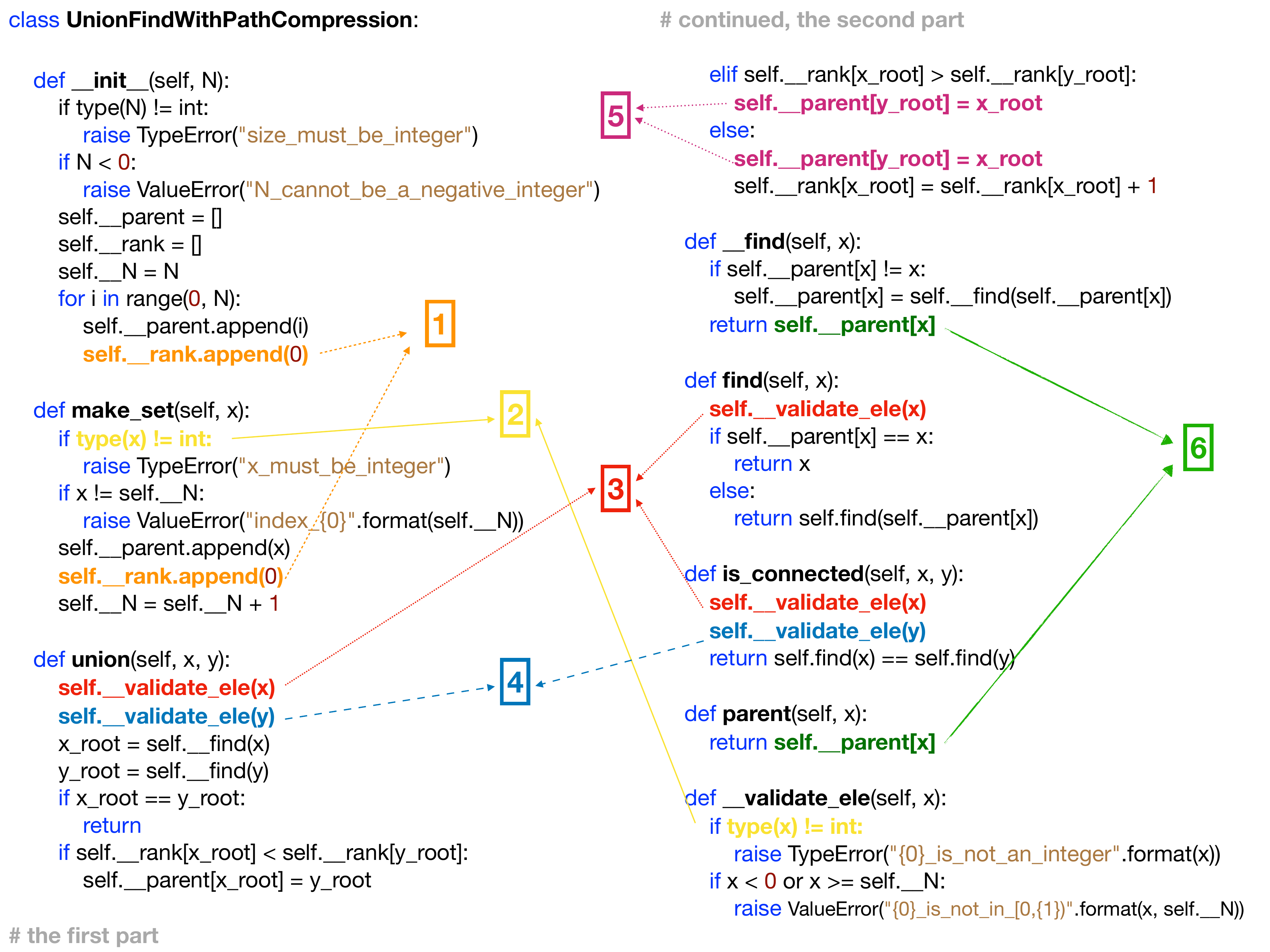}
    \end{minipage}
}
\vspace{-10pt}
\caption{Examples that {\our} Fail to Handle. Left: Sieve of Atkin Algorithm; Right: Union Find Class Class. (The duplicated contents are highlighted in colored bolded font).}\label{fig:failed}
\vspace{-10pt}
\end{figure*}

\subsubsection{Program Completion with Code Fragments}

Table~\ref{tab:completion} covers the program code completion experimental results of the comparison methods. Here, for each program in the dataset, we randomly pick one line in the program as the input for the models to complete the program code (i.e., generate the code ahead of or after the input line). Slightly different from the program generation as shown in Table~\ref{tab:generation_complete}, where the input is the tree root node content, the input in the program completion can be any lines in the program code, results obtained in which are slightly lower than those in Table~\ref{tab:generation_complete}.

Among these $43$ programs in the dataset, {\our} completes $34$ of the correctly, which is much better than the other baseline methods. The AST-BiRNN methods can still complete about $10$ of the programs correctly, while the remaining methods cannot complete the program code at all. The program completion task may require the model to be able to generate contents in both the sequential directions and the hierarchical directions, i.e., ahead of the input, after the input, above the input and below the input, which can demonstrate the advantages of the {\unit} model compared against the traditional RNN models.

\subsubsection{Experimental Discoveries}

Generally, according to the program generation, program interpretation and program completion experimental results, {\our} performs very well in inferring both the contents at both the children nodes and father node. The main reason can be due to that the {\unit} model effectively captures both the sequential and hierarchical information patterns in the ASTs, which performs much more effectively than the models merely capturing the sequential patterns. In addition, the AST will also greatly improve the model performance, since the tree diagram will effectively help the models outline the program hierarchical structure. The {\autoencoder} model cannot achieve a good performance in these content generation tasks.

\subsection{Case Study}
In this part, we will provide a study about the succeeded and failed cases of {\our} in the experiments.

\subsubsection{Succeeded Cases}

In Figures~\ref{fig:succeeded}, we show the program code that {\our} can handle very well in both generation, interpretation and completion. The left program code is about the Binary Search algorithm and the right code is about the Quick Sort algorithm. We show both the original and the generated program code of both algorithms in the plots. By checking these two program code blocks, we can observe that the code generated by {\our} can implement exactly the same function as the original program code in the dataset. Furthermore, we can also observe some differences between the program code blocks: (1) many of the operator and variable tokens in the original code blocks are connected, while the tokens in the generated code block are well separated; (2) in the generated code block, some extra parentheses are inserted, especially for some expressions in statements; (3) the code indent in the original code uses two space keys, but in the generated code involves 1 tab key instead. These differences are mainly due to our model {\our} is trained based on the AST, whose contents are well organized and structured by the program parser. 

\subsubsection{Failed Cases}

Besides the succeeded cases aforementioned, in Figure~\ref{fig:failed}, we also provide two cases that {\our} cannot handle well, especially in program code generation and completion. The left program code is about the Sieve of Atkin algorithm, and the right code is about the Union Find class with compressed path. The main problem with the code is that it contains so many duplicated contents. In the blocks, we can identify several common statements (in the same colors), which will make the {\our} fail to work. For instance, in the Union Find class code, given a statement ``self.\_\_validate\_ele(y)'' (in bolded blue font), it will be very hard for the model to generate the statement after it, since there are two different options ``x\_root = self.\_\_find(x)'' and ``return self.find(x) == self.find(y)''. Such a problem can be hopefully addressed by incorporate a even deeper architecture in {\our}. Just like this failed case, if we can effectively incorporate the father node of ``self.\_\_validate\_ele(y)'' (i.e., ``def union(self, x, y):'' and 	``def is\_connected(self, x, y):''), the conflict can be resolved promisingly. We will leave it as a potential future work.

\section{Related Work} \label{sec:related_work}

The problem studied in this paper is strongly correlated with research problems about \textit{deep neural network}, \textit{text generation} and \textit{program synthesis}. 

\vspace{5pt}

\noindent \textbf{Deep Neural Networks}: The essence of deep learning is to compute hierarchical features or representations of the observational data \cite{GBC16, LBH15}. With the surge of deep learning research and applications in recent years, lots of research works have appeared to apply the deep learning methods, like deep belief network \cite{HOT06}, deep Boltzmann machine \cite{SH09}, Deep neural network \cite{J02, KSH12} and Deep autoencoder model \cite{VLLBM10}, in various applications, like speech and audio processing \cite{DHK13, HDYDMJSVNSK12}, language modeling and processing \cite{ASKR12, MH09}, information retrieval \cite{H12, SH09}, objective recognition and computer vision \cite{LBH15}, as well as multimodal and multi-task learning \cite{WBU10, WBU11}.

\vspace{5pt}

\noindent \textbf{Text Generation}: Text generation has been an important problem in both text mining and natural language processing. Depending on the input information, the text generation problem can be categorized into text generation from \textit{keywords} \cite{UIS02}, \textit{concepts} \cite{KL13}, \textit{topics} \cite{CH15}, \textit{ontologies} \cite{B05} and \textit{images} \cite{VTBE14}. In \cite{UIS02}, the authors propose a method consisting of candidate-text construction and evaluation for sentence generation from keywords/headwords. Konstas et al. \cite{KL13} introduced a global model for concept-to-text generation, which refers to the task of automatically producing textual output from non-linguistic input. In terms of the objective output, the text generation problem includes \textit{question generation} \cite{CH15}, \textit{image captions} \cite{VTBE14} and \textit{image descriptions} \cite{KPDLCBB11}. Various models have been used in the text generation problems, including \textit{RNN} \cite{SMH11}, \textit{Autoencoder} \cite{LLJ15} and \textit{GAN} \cite{ZGFCHSC17}.

\vspace{5pt}

\noindent \textbf{Program Synthesis}: The problem studied in this paper is also closely related with the \textit{program synthesis} problem studied in software engineering. Formally, the goal of software program synthesis is to generate programs automatically from high-level specifications, lots of research works have been done on this topic already. Program synthesis is a challenging problem, which may require external supervisions from either \textit{template} \cite{SGF13}, \textit{examples} and \textit{type information} \cite{OZ15}, and \textit{oracles} \cite{JGST10}. In \cite{JGST10}, the authors present a novel approach to automatic synthesis of loop-free programs based on a combination of oracle-guided learning from examples. Based on the program templates, \cite{SGF13} introduces an approach to generate the programs from the templates. Osera et al. \cite{OZ15} introduce an algorithm for synthesizing recursive functions that process algebraic datatypes, which exploits both type information and input-output examples to prune the search space. Some other program synthesis works address the problem with \textit{recursive algorithms} \cite{AGK13}, \textit{deductive approach} \cite{MW80}, \textit{crowd-sourcing} \cite{CDLMV15}, and \textit{program verification} \cite{SGF10}.
\section{Conclusion}\label{sec:conclusion}

In this paper, we have studied a novel research problem about program generation, which covers the tasks about \textit{program code generation}, \textit{program interpretation} and \textit{program completion}. To address such a challenging problem, a new learning model, namely {\our}, has been introduced. {\our} learns the program code contents by parsing the program code into ASTs, whose nodes contain the program code/comment contents while the AST structure can indicate the program logic flows. To effectively capture both the hierarchical and sequential patterns in the ASTs, a new unit model, i.e., {\unit}, is introduced as the basic structure covered in {\our}. We have tested the effectiveness of {\our} on real-world program datasets, and {\our} can achieve very outstanding performance than the other state-of-the-art baseline methods in addressing the program generation tasks.
\balance
\bibliographystyle{plain}
\bibliography{12_reference}

\begin{thebibliography}{10}

\bibitem{codebase}
Codebases: Millions of lines of code.
\newblock
  \url{http://www.informationisbeautiful.net/visualizations/million-lines-of-code/}.
\newblock [Online; accessed 2-December-2017].

\bibitem{grow}
Engineering software market to reach \$50.34 bn in 2022.
\newblock
  \url{https://www.automation.com/automation-news/industry/engineering-software-market-to-reach-5034-bn-in-2022}.
\newblock [Online; accessed 2-December-2017].

\bibitem{market}
The labor market supply \& demand of software developers.
\newblock
  \url{http://www.economicmodeling.com/2017/06/01/labor-market-supply-demand-software-developers/}.
\newblock [Online; accessed 2-December-2017].

\bibitem{google}
Why google stores billions of lines of code in a single repository.
\newblock
  \url{https://cacm.acm.org/magazines/2016/7/204032-why-google-stores-billions-of-lines-of-code-in-a-single-repository/fulltext}.
\newblock [Online; accessed 2-December-2017].

\bibitem{AGK13}
A.~Albarghouthi, S.~Gulwani, and Z.~Kincaid.
\newblock Recursive program synthesis.
\newblock In {\em CAV}, 2013.

\bibitem{ASKR12}
E.~Arisoy, T.~Sainath, B.~Kingsbury, and B.~Ramabhadran.
\newblock Deep neural network language models.
\newblock In {\em WLM}, 2012.

\bibitem{B05}
K.~Bontcheva.
\newblock Generating tailored textual summaries from ontologies, 2005.

\bibitem{CH15}
Y.~Chali and S.~Hasan.
\newblock Towards topic-to-question generation.
\newblock {\em Comput. Linguist.}, 2015.

\bibitem{CGCB14}
J.~Chung, C.~Gulcehre, K.~Cho, and Y.~Bengio.
\newblock {\em Empirical evaluation of gated recurrent neural networks on
  sequence modeling}.
\newblock 2014.

\bibitem{CDLMV15}
R.~Cochran, L.~DAntoni, B.~Livshits, D.~Molnar, and M.~Veanes.
\newblock Program boosting: Program synthesis via crowd-sourcing.
\newblock In {\em POPL}, 2015.

\bibitem{DHK13}
L.~Deng, G.~Hinton, and B.~Kingsbury.
\newblock New types of deep neural network learning for speech recognition and
  related applications: An overview.
\newblock In {\em ICASSP}, 2013.

\bibitem{GBC16}
I.~Goodfellow, Y.~Bengio, and A.~Courville.
\newblock {\em Deep Learning}.
\newblock MIT Press, 2016.
\newblock \url{http://www.deeplearningbook.org}.

\bibitem{H12}
S.~Hill.
\newblock Elite and upper-class families.
\newblock In {\em Families: A Social Class Perspective}. 2012.

\bibitem{HDYDMJSVNSK12}
G.~Hinton, L.~Deng, D.~Yu, G.~Dahl, A.~Mohamed, N.~Jaitly, A.~Senior,
  V.~Vanhoucke, P.~Nguyen, T.~Sainath, and B.~Kingsbury.
\newblock Deep neural networks for acoustic modeling in speech recognition.
\newblock {\em IEEE Signal Processing Magazine}, 2012.

\bibitem{HOT06}
G.~Hinton, S.~Osindero, and Y.~Teh.
\newblock A fast learning algorithm for deep belief nets.
\newblock {\em Neural Comput.}, 2006.

\bibitem{HS97}
S.~Hochreiter and J~Schmidhuber.
\newblock Long short-term memory.
\newblock {\em Neural Comput.}, 1997.

\bibitem{J02}
H.~Jaeger.
\newblock {Tutorial on training recurrent neural networks, covering BPPT, RTRL,
  EKF and the ``echo state network'' approach}.
\newblock Technical report, Fraunhofer Institute for Autonomous Intelligent
  Systems (AIS), 2002.

\bibitem{JGST10}
S.~Jha, S.~Gulwani, S.~Seshia, and A.~Tiwari.
\newblock Oracle-guided component-based program synthesis.
\newblock In {\em ICSE}, 2010.

\bibitem{KL13}
I.~Konstas and M.~Lapata.
\newblock A global model for concept-to-text generation.
\newblock {\em J. Artif. Int. Res.}, 2013.

\bibitem{KSH12}
A.~Krizhevsky, I.~Sutskever, and G.~Hinton.
\newblock Imagenet classification with deep convolutional neural networks.
\newblock In {\em NIPS}, 2012.

\bibitem{KPDLCBB11}
G.~Kulkarni, V.~Premraj, S.~Dhar, S.~Li, Y.~Choi, A.~Berg, and T.~Berg.
\newblock Baby talk: Understanding and generating simple image descriptions.
\newblock In {\em CVPR}, 2011.

\bibitem{LBH15}
Y.~LeCun, Y.~Bengio, and G.~Hinton.
\newblock {Deep learning}.
\newblock {\em Nature}, 521, 2015.
\newblock \url{http://dx.doi.org/10.1038/nature14539}.

\bibitem{LLJ15}
J.~Li, M.~Luong, and D.~Jurafsky.
\newblock A hierarchical neural autoencoder for paragraphs and documents.
\newblock In {\em ACL}, 2015.

\bibitem{MW80}
Z.~Manna and R.~Waldinger.
\newblock A deductive approach to program synthesis.
\newblock {\em ACM Trans. Program. Lang. Syst.}, 1980.

\bibitem{MH09}
A.~Mnih and G.~Hinton.
\newblock A scalable hierarchical distributed language model.
\newblock In {\em NIPS}. 2009.

\bibitem{OZ15}
P.~Osera and S.~Zdancewic.
\newblock Type-and-example-directed program synthesis.
\newblock In {\em PLDI}, 2015.

\bibitem{SH09}
R.~Salakhutdinov and G.~Hinton.
\newblock Semantic hashing.
\newblock {\em International Journal of Approximate Reasoning}, 2009.

\bibitem{SP97}
M.~Schuster and K.K. Paliwal.
\newblock Bidirectional recurrent neural networks.
\newblock {\em Trans. Sig. Proc.}, 1997.

\bibitem{SGF10}
S.~Srivastava, S.~Gulwani, and J.~Foster.
\newblock From program verification to program synthesis.
\newblock In {\em POPL}, 2010.

\bibitem{SGF13}
S.~Srivastava, S.~Gulwani, and J.~Foster.
\newblock Template-based program verification and program synthesis.
\newblock {\em International Journal on Software Tools for Technology
  Transfer}, 2013.

\bibitem{SMH11}
I.~Sutskever, J.~Martens, and G.~Hinton.
\newblock Generating text with recurrent neural networks.
\newblock In {\em ICML}, 2011.

\bibitem{UIS02}
K.~Uchimoto, H.~Isahara, and S.~Sekine.
\newblock Text generation from keywords.
\newblock In {\em COLING}, 2002.

\bibitem{VLLBM10}
P.~Vincent, H.~Larochelle, I.~Lajoie, Y.~Bengio, and P.~Manzagol.
\newblock Stacked denoising autoencoders: Learning useful representations in a
  deep network with a local denoising criterion.
\newblock {\em Journal of Machine Learning Research}, 2010.

\bibitem{VTBE14}
O.~Vinyals, A.~Toshev, S.~Bengio, and D.~Erhan.
\newblock Show and tell: A neural image caption generator, 2014.

\bibitem{WBU10}
J.~Weston, S.~Bengio, and N.~Usunier.
\newblock Large scale image annotation: Learning to rank with joint word-image
  embeddings.
\newblock {\em Journal of Machine Learning}, 2010.

\bibitem{WBU11}
J.~Weston, S.~Bengio, and N.~Usunier.
\newblock Wsabie: Scaling up to large vocabulary image annotation.
\newblock In {\em IJCAI}, 2011.

\bibitem{ZGFCHSC17}
Y.~Zhang, Z.~Gan, K.~Fan, Z.~Chen, R.~Henao, D.~Shen, and L.~Carin.
\newblock Adversarial feature matching for text generation.
\newblock {\em arXiv preprint arXiv:1706.03850}, 2017.

\end{thebibliography}

\end{document}